\documentclass{article}
\usepackage{ijcai26}
\usepackage{times}
\usepackage{soul}
\usepackage{url}
\usepackage[hidelinks]{hyperref}
\usepackage[utf8]{inputenc}
\usepackage[small]{caption}
\usepackage{graphicx}
\usepackage{tikz}
\usetikzlibrary{arrows.meta, positioning, shapes.geometric, calc}
\usepackage{amsmath}
\usepackage{booktabs}
\usepackage[switch]{lineno}

\title{Policy as Code: A Coroutine-Bridge Harness for\\ Fast-Reasoning Reliability on CAR-bench}

\author{
    Ivan Matveev
    \affiliations
    Proxima Ultra
    \emails
    i.a.matveev@gmail.com
}

\begin{document}

\maketitle

\begin{abstract}
CAR-bench evaluates whether tool-using agents stay reliable under real-world
uncertainty, executing every tool inside the evaluator so that each tool-result
exchange is a separate agent round-trip. A conventional next-action agent can
batch parallel tool calls, but a chain of dependent calls costs it one model
call per round of results. We present a coroutine-bridge
harness in which the model's only action is to emit a Python program that
blocks and resumes \emph{in place} across evaluator tool exchanges. This
decouples model invocation from tool round-trips: on the public test split the
agent uses a median of two model calls against seven agent turns per task,
resolving a full multi-turn task in a median of 1.8\,s of model latency on
Cerebras \texttt{gpt-oss-120b}. Because the action surface is executable code,
deterministic CAR-bench policies are encoded directly as logic in the tool
layer rather than as prompt rules, enforcing compliance at zero reasoning cost.
On the official hidden evaluation the harness won Track~2 with 60.0\%
Pass\textsuperscript{3}, $4.5\times$ the organizer baseline, at the lowest
estimated cost and the fastest median task latency (3.14\,s) of any entry
scoring above that baseline; the same unchanged harness reproduced an identical
60.0\% Pass\textsuperscript{3} on GPT-5.5 in the Open track, matching
frontier-model agents. A single static prompt, appended with per-task
state at the tail, stays byte-identical across calls and across tasks: the
frozen submission prompt served 78\% of input tokens from cache (86.6\% across
its warm tail), against 73\% over a three-week development corpus in which
prompt edits repeatedly reset the cache. This compounds the few-call design
into a small fraction of nominal input compute.
\end{abstract}

\section{Introduction}

CAR-bench measures the \emph{consistency} and \emph{limit-awareness} of LLM
agents in an in-car assistant domain of 58 interconnected tools, 19 domain
policies, and 254 public tasks spanning base, hallucination, and
disambiguation categories~\cite{kirmayr-etal-2026-car}. The target is
deployment reliability across repeated trials: the headline metric is
Pass\textsuperscript{3}, which credits a task only when it passes all three
independent trials.

Track~2 restricts the model to Cerebras-hosted \texttt{gpt-oss} and rewards
\emph{compute-aware} harnessing: architectures that convert fast inference into
low end-to-end latency and reliable behavior. A defining property of the
benchmark shapes what ``compute-aware'' means here. Unlike a standard
Agent-to-Agent (A2A) deployment, where an agent executes its own tools
privately and one request spans the whole task~\cite{a2a2025}, CAR-bench
\emph{externalizes} tool execution to the evaluator so it can score the tool
trajectory and inject environment perturbations. Each batch of tool calls the
agent emits returns as a fresh A2A message carrying tool results. Batching makes
independent calls cheap for any agent, but a chain of \emph{dependent} calls ---
where the next call needs the previous result --- is different: a conventional
``next-action'' agent, which the reference templates describe as one model call
per assistant step, must spend a model call on each round of results to decide
the next call.

Our contribution is a harness that removes this tax. The model emits a single
executable Python program~\cite{wang2024codeact}; when that program calls a
tool wrapper, a coroutine bridge suspends the Python thread, emits the official
tool call, and resumes the \emph{same} stack frame when results arrive ---
without a new model call. One program can thus drive many sequential and
parallel tool round-trips per model invocation. Two consequences follow, and
both are the substance of this report: (i)~model calls, and hence model latency
on the critical path, drop sharply relative to agent turns; and (ii)~because the
action surface is code, deterministic policy is expressed as ordinary logic in
the tools the code calls, so policy compliance costs the model no attention and
no extra reasoning turns.

\section{The Coroutine Bridge}

\subsection{Externalized Tools Force Round-Trips}

In CAR-bench the participant agent never executes vehicle, navigation, weather,
or productivity tools. It returns a data part of \texttt{tool\_calls}; the
evaluator executes them, advances environment state, and returns a new A2A
message of \texttt{tool\_results}. This is what enables trajectory scoring and
hallucination injection, and it is why the benchmark cannot be treated as a
private single-request task. The cost is structural: an $n$-step dependent tool
chain is $n$ separate agent round-trips.

\subsection{Decoupling Model Calls from Tool Round-Trips}

Each A2A \texttt{context\_id} owns a long-lived worker holding one persistent
Python interpreter, a tool bridge, a scratchpad, and a compact transcript. The
model's sole action is \texttt{execute\_python}. Execution proceeds as follows:

\begin{enumerate}
\itemsep2pt
\item The model emits one Python program (one model call).
\item The program runs in the worker thread until it calls a tool wrapper such
  as \texttt{get\_current\_navigation\_state(...)}.
\item The bridge places the official tool call on the outbox and \emph{blocks}
  the Python thread.
\item The evaluator executes the tool and returns a tool-results message.
\item The bridge delivers those results directly into the blocked call, and the
  same stack frame resumes --- \emph{no model call}.
\item The program branches, calls further tools, or emits the user-facing
  response.
\end{enumerate}

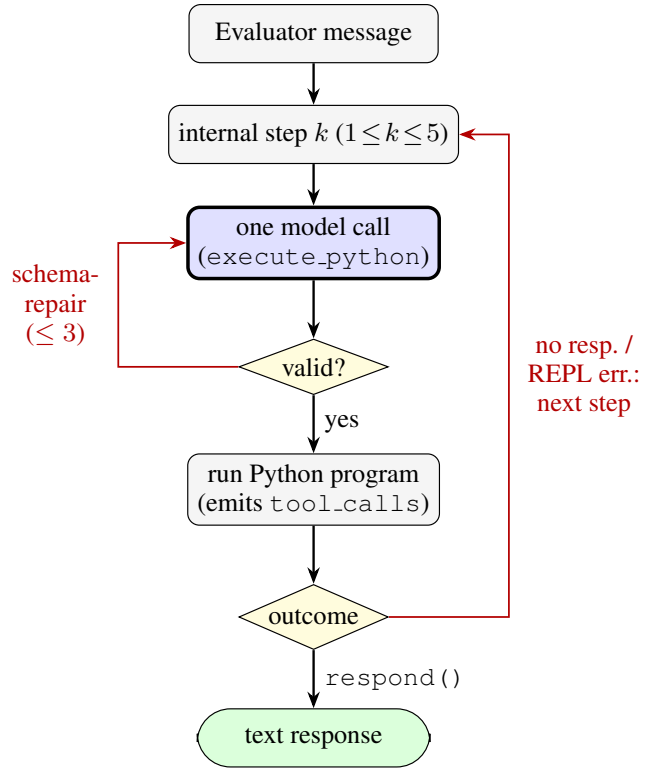
\begin{figure}[t]
\centering
\resizebox{\columnwidth}{!}{%
\begin{tikzpicture}[
  font=\small,
  proc/.style={draw, rounded corners, minimum height=7mm, minimum width=30mm,
               align=center, fill=gray!8},
  call/.style={draw, rounded corners, minimum height=7mm, minimum width=30mm,
               align=center, fill=blue!12, very thick},
  dec/.style={draw, diamond, aspect=2.2, inner sep=1pt, minimum width=18mm,
              align=center, fill=yellow!16},
  term/.style={draw, rounded corners=4mm, minimum height=7mm, minimum width=28mm,
               align=center, fill=green!14},
  a/.style={-{Stealth[length=2mm]}, thick},
  loop/.style={-{Stealth[length=2mm]}, semithick, red!70!black},
  llbl/.style={font=\footnotesize, red!70!black, align=center},
]
\node[proc] (in) {Evaluator message};
\node[proc, below=5mm of in] (step) {internal step $k$ ($1\!\le\!k\!\le\!5$)};
\node[call, below=5mm of step] (model) {one model call\\(\texttt{execute\_python})};
\node[dec, below=7mm of model] (valid) {valid?};
\node[proc, below=7mm of valid] (run) {run Python program\\{\footnotesize(emits \texttt{tool\_calls})}};
\node[dec, below=7mm of run] (out) {outcome};
\node[term, below=7mm of out] (done) {text response};
\draw[a] (in) -- (step);
\draw[a] (step) -- (model);
\draw[a] (model) -- (valid);
\draw[a] (valid) -- node[right]{yes} (run);
\draw[a] (run) -- (out);
\draw[a] (out) -- node[right]{\texttt{respond()}} (done);
\coordinate (LX) at ($(step.west)+(-0.6,0)$);
\coordinate (RX) at ($(step.east)+(0.6,0)$);
\draw[loop] (valid.west) -- (valid.west -| LX) -- (model.west -| LX) -- (model.west);
\coordinate (Lmid) at ($(valid.west)!0.5!(model.west)$);
\node[llbl, anchor=east] at ([xshift=-3pt]Lmid -| LX) {schema-\\repair\\($\le3$)};
\draw[loop] (out.east) -- (out.east -| RX) -- (step.east -| RX) -- (step.east);
\coordinate (Rmid) at ($(out.east)!0.5!(step.east)$);
\node[llbl, anchor=west] at ([xshift=3pt]Rmid -| RX) {no resp.\ /\\REPL err.:\\next step};
\end{tikzpicture}}
\caption{Per-user-turn call-depth audit. The worker issues at most five
sequential model calls (the $k\le5$ loop); each step is one
\texttt{execute\_python} call, so there is \emph{no} parallel LLM fan-out.
Inside a program the model branches over tool results (the diamonds) and issues
parallel \emph{tool} calls via \texttt{batch(...)}, which resume the blocked
Python frame with no model call and do not count toward the limit. Two retry
loops --- schema-repair ($\le3$) and provider backoff ($\le4$) --- recover a
malformed or failed call rather than add depth. Per baseline step the sequential
LLM-call depth is thus $\le5$ and typically below one; measured median two calls
per task, p90 four.}
\label{fig:audit}
\end{figure}

The model is invoked once per \emph{decision}: one program spans many tool
results before the next call (Figure~\ref{fig:audit} audits this loop).
Crucially, this hides nothing: every tool call still appears in the official
A2A trajectory exactly as the evaluator expects, so the technique is fully
compliant. What changes is only the number of \emph{model} invocations between
tool round-trips. Table~\ref{tab:compute} quantifies the effect: a median of
two model calls carries a median of seven agent turns per task.

\paragraph{Unconstrained action format.}
The model emits its action as a plain-text JSON object --- a short thought and a
Python program --- which the runtime parses, rather than forcing validity
through grammar-constrained or native structured decoding. This follows evidence
that constrained decoding trades a small but measurable accuracy cost for token
savings~\cite{matveev2026toon,schall2025structure}; malformed actions are caught
instead by a bounded schema-repair retry.

\subsection{Parallel Batching}

Independent reads are issued together via \texttt{batch([...])}, which emits one
parallel A2A tool-call message and returns results in input order. Dependent
calls remain ordinary sequential Python. Batching collapses independent tool
round-trips in addition to the model-call savings, further shortening the
critical path without sacrificing the natural imperative control flow that
makes code actions expressive~\cite{wang2024codeact}.

\section{Policy as Code}

Many CAR-bench policies are \emph{deterministic given grounded facts}: whether
opening a window above 25\% requires an AC confirmation, whether high beams are
blocked while fog lights are on, which route default applies to a newly created
segment. The CodeAct action surface lets us encode such policy directly as
logic in the tool layer rather than as prose the model must recall. This is the
central design principle of the harness.

We separate two responsibilities. The \emph{runtime} owns facts and strict
policy; the \emph{model} owns interpretation of the user's intent. Concretely,
policy-critical raw wrappers delegate to guarded implementations
(\texttt{set\_fog\_lights} to a fog-light safety helper,
\texttt{open\_close\_window} to an AC-aware helper), so the policy executes
whether the model calls the public name or the helper. A confirmation state
machine stores the fully grounded pending action and resumes the exact call on
the user's next turn. Result normalization gives the model stable field names
so a simple, direct tool path is not silently trapped by a dynamic result key.
Missing capabilities are resolved reactively against the live per-task tool
surface --- a membership test the organizers confirmed as permitted --- never
by comparing against a bundled catalog.

The practical test for each policy is: \emph{is it deterministic given grounded
facts?} If yes, it becomes enforced code in a tool; if it requires genuine
interpretation of the request, it stays in the prompt. A policy expressed only
in the prompt can be forgotten under attention pressure; a policy expressed as
tool code is enforced by construction and, on the critical path, free.

\section{Reliability Mechanisms}

Three further mechanisms address the specific CAR-bench failure classes.
\textbf{Response obligations}: policy-mandated disclosures (e.g.\ a toll-road
mention, a $>$3\textdegree C cross-zone temperature warning) are registered as
obligations that the response step appends only if the model's own text omitted
them, preserving compound-request flexibility without making warnings optional.
\textbf{Unknown-value sentinels}: hallucination tasks can blank fields in
otherwise successful results; the string \texttt{"unknown"} is normalized to a
sentinel that is storage- and serialization-safe but aborts with a
missing-response acknowledgement the moment model code tries to use it in a
meaningful operation, converting a silent fabrication into an honest limitation.
\textbf{Turn-scoped outcome tracking}: mutation failures are keyed by tool and
target arguments and survive every Python block in a user turn, so an optimistic
success claim cannot outrun a failed side effect.

\section{Results}

We report the organizers' hidden-set evaluation (30 tasks, 3 trials, scored by
the organizers) and our own development evaluation on the public \emph{test}
split (125 tasks, 3 trials $=$ 375 evaluations). Both use Cerebras
\texttt{gpt-oss-120b} at temperature~0 with a JSON code-action contract and the
organizers' default Gemini~2.5~Flash simulator and judge.
Table~\ref{tab:official} gives the official hidden-set result and
Table~\ref{tab:reliability} the per-category public-split breakdown; the hidden
set is markedly harder than the public one.

\begin{table}[t]
\centering
\begin{tabular}{lrrr}
\toprule
Per task (test, 3 trials) & Median & Mean & p90 \\
\midrule
Model calls          & 2      & 2.4   & 4     \\
Agent (A2A) turns    & 7      & 8.0   & 12    \\
Model latency (s)    & 1.8    & 2.4   & 4.4   \\
Output tokens        & 1{,}443 & 1{,}850 & 3{,}625 \\
Input tokens         & 78.2k  & 87.8k & 165.6k \\
\bottomrule
\end{tabular}
\caption{Compute and latency profile. Two model calls carry seven agent turns:
the coroutine bridge decouples model invocation from evaluator tool
round-trips. Input token budget is 500k/task; observed mean is 88k.}
\label{tab:compute}
\end{table}

\begin{table}[t]
\centering
\small
\begin{tabular}{lrrr}
\toprule
Hidden set (Track~2) & Ours & 2nd & Baseline \\
\midrule
Pass\textsuperscript{3} (\%)      & \textbf{60.00} & 50.00 & 13.33 \\
Pass@3 (\%)                       & 66.67 & 76.67 & 43.33 \\
Successful trials                 & 57/90 & 57/90 & 26/90 \\
Success consistency (\%)          & \textbf{92.50} & 73.91 & 50.00 \\
Task latency, median (s)          & \textbf{3.14} & 9.82 & 5.03 \\
Mean tokens / trial               & 112{,}993 & 342{,}300 & 140{,}956 \\
Est.\ cost / trial (\$)           & \textbf{0.041} & 0.13 & 0.053 \\
\bottomrule
\end{tabular}
\caption{Official hidden-set result: first of 13 Track~2 submissions, against
the runner-up and the non-competing organizer baseline. Our entry and the
runner-up convert the \emph{same} 57/90 successful trials into very different
Pass\textsuperscript{3} (60.0 vs 50.0) because consistency differs (92.5 vs
73.9).}
\label{tab:official}
\end{table}

\begin{table}[t]
\centering
\small
\begin{tabular}{lrr}
\toprule
Public test split & Pass\textsuperscript{1} & Pass\textsuperscript{3} \\
\midrule
Base            & 90.0 & 78.0 \\
Hallucination   & 98.7 & 96.0 \\
Disambiguation  & 90.7 & 88.0 \\
\midrule
Overall         & 93.3 & 87.3 \\
\bottomrule
\end{tabular}
\caption{Per-category reliability on the public test split (\%; 125 tasks, 3
trials). Hallucination tasks --- acknowledging a missing capability instead of
fabricating one --- are the strongest category, which the runtime's live
tool-surface checks and unknown-value sentinels are built to enforce.}
\label{tab:reliability}
\end{table}

\paragraph{Latency.}
A complete multi-turn task resolves in a median of 1.8\,s of model latency
(Table~\ref{tab:compute}). We report \emph{model} latency deliberately: of the
5{,}351\,s wall time for the full 375-evaluation run, only 899\,s was
agent LLM time; the remainder is the evaluator's own simulator and judge, which
are not attributable to the agent. Fast per-call inference on Cerebras is thus
compounded by a harness that issues few calls.

\paragraph{Reliability: consistency is what decides
Pass\textsuperscript{3}.}
The hidden-set result isolates what the harness actually buys
(Table~\ref{tab:official}). Our entry and the runner-up recorded the
\emph{identical} 57/90 successful trials, yet scored 60.0\% and 50.0\%
Pass\textsuperscript{3}. The runner-up in fact solved \emph{more} tasks at least
once (Pass@3 76.7 vs 66.7); it converted fewer of them into three-for-three
passes (success consistency 73.9\% vs 92.5\%). Because deterministic policy is
enforced in tool code rather than re-derived by the model on each trial, a task
we solve once is very likely to be solved every time --- exactly the property
Pass\textsuperscript{3} rewards. Against the organizer baseline on the same
model, Pass\textsuperscript{3} rises $4.5\times$ (60.0 vs 13.3) while median
latency drops 38\% and mean tokens per trial fall 20\%.

\paragraph{Cross-model reproduction.}
Submitted unchanged to the Open track on GPT-5.5, the same harness scored the
same 60.0\% Pass\textsuperscript{3} (tied 4th--6th of 21) with the lowest median
task latency in that track (14.1\,s). The decomposition differs, however: the
frontier model covers more tasks (Pass@3 76.7\% vs 66.7\%; 60/90 vs 57/90
successful trials) and converts fewer of them (success consistency 80.4\% vs
92.5\%). Model capacity buys coverage; the harness buys consistency; on this
hidden set the two effects cancel to the same Pass\textsuperscript{3}. The rise
in coverage also bounds a concern that the identical headline score would
otherwise raise, since a harness that capped the stronger model would have held
its coverage down.

\paragraph{Prefix-cache efficiency.}
The system prompt is one large static block --- runtime rules, the domain skill,
and the full tool catalog --- and per-task state is appended only at the tail, so
the cacheable prefix is byte-identical across every model call and, crucially,
\emph{across tasks}. We deliberately do not swap skills per task, which would
break this prefix. Once the cache is warm, per-minute input-token cache rates
run to \textbf{88.7\% (p90) and 99.9\% at peak}, and a stable-prompt test day
held 86.6\% over its final half hour. Across the entire development corpus (918M
input tokens over three weeks) the token-weighted rate is 73.2\% (672M of 918M),
held down by the frequent prompt edits that reset the cache during active
development; the deployed agent runs a frozen prompt, so a long hidden-set
evaluation runs warm. Combined with the two-call-per-task design, real per-task
input compute is a small fraction of the nominal $\sim$88k tokens. The mechanism
is provider-general: any prefix-caching backend (Cerebras here, and the OpenAI
route used in our Open-track submission) benefits from the same stable-prefix,
tail-accumulation structure.

\paragraph{Compute compliance.}
Figure~\ref{fig:audit} audits the call structure of one user turn. The worker
issues at most five sequential model calls (the $k\le5$ internal-step cap), each
a single call, so there is no parallel LLM fan-out; parallel \emph{tool} calls
via \texttt{batch(...)} resume the blocked Python frame without a model call and
do not count toward the limit. Two bounded retry loops --- schema-repair
($\le3$ attempts) and provider backoff ($\le4$) --- recover a malformed or failed
call rather than add reasoning depth. Observed usage is a median of two model
calls per task (p90 four). Mean input+output stays under 90k tokens against the
500k budget. Token and latency accounting is reported through
\texttt{turn\_metrics} on each concluding response; because the accumulator sums
across intermediate tool exchanges and the evaluator records metrics only on
non-tool-call responses, per-task totals are complete.

\paragraph{Development discipline.}
Two rules kept the harness from overfitting the weaker model. First, a change
was accepted only if a stronger model already outperformed a weaker one on the
bare prompt, so gains reflect structure rather than scaffolding. Second, a
persistent \texttt{gpt-oss-120b} failure was first reproduced on a stronger
model: if the stronger model passed, the ceiling was reachable and the fix was
to encode the missing determinism as tool code (Policy as Code); if both
failed, the defect lay in existing harness and was repaired there rather than
patched over. Candidate changes were judged on consistent-failure sets across
runs, never single noisy trials.

\section{Limitations and Conclusion}

Two limits are visible in the official numbers. First, 40\% of hidden tasks
still fail at least one trial: the residual gap comes from simulator and judge
variance and from rare paths where a temperature-0 model chooses a
valid-looking but wrong action despite enforced structure. The GPT-5.5 run
locates that ceiling more precisely: a stronger model gains ten points of
coverage and gives back twelve of consistency, so the residual sits in what the
harness has been taught to make deterministic rather than in what the model can
work out unaided. Second, our Pass@3 (66.7\%) is lower than the runner-up's (76.7\%):
enforcing one grounded path yields consistency, and it also forecloses lucky
recoveries that a more exploratory agent sometimes finds. Raising Pass@3
without spending the consistency that Pass\textsuperscript{3} rewards is the
open problem this design leaves. Worker state is also in memory and not
recoverable across process restarts.

The coroutine bridge shows that fast inference under CAR-bench's
externalized-tool design is best exploited by letting each fast call accomplish
an entire branching program, with deterministic policy compiled into the tools
that program invokes. On the hidden evaluation that combination won Track~2 at
the lowest cost and fastest latency above baseline, and reproduced its score on
a frontier model without a single architectural change.

\bibliographystyle{named}
\bibliography{track2_report}

\end{document}